\documentclass{Interspeech}

\interspeechcameraready

\title{Investigating Affect Mining Techniques for Annotation Sample Selection in the Creation of Finnish Affective Speech Corpus}

\author[affiliation={1}]{Kalle}{Lahtinen}
\author[affiliation={1}]{Einari}{Vaaras}
\author[affiliation={2}]{Liisa}{Mustanoja}
\author[affiliation={1}]{Okko}{Räsänen}

\affiliation{Signal Processing Research Centre}{Tampere University}{Finland}
\affiliation{Language Studies}{Tampere University}{Finland}
\email{firstname.surname@tuni.fi}

\usepackage{comment}
\usepackage{xcolor}%
\usepackage{makecell}
\usepackage{multirow}
\usepackage{microtype}

\usepackage[noadjust]{cite}

\begin{document}

\maketitle

\begin{abstract}
    
    Study of affect in speech requires suitable data, as emotional expression and perception vary across languages. Until now, no corpus has existed for natural expression of affect in spontaneous Finnish, existing data being acted or from a very specific communicative setting. This paper presents the first such corpus, created by annotating 12,000 utterances for emotional arousal and valence, sampled from three large-scale Finnish speech corpora. To ensure diverse affective expression, sample selection was conducted with an affect mining approach combining acoustic, cross-linguistic speech emotion, and text sentiment features. We compare this method to random sampling in terms of annotation diversity, and conduct post-hoc analyses to identify sampling choices that would have maximized the diversity. As an outcome, the work introduces a spontaneous Finnish affective speech corpus and informs sampling strategies for affective speech corpus creation in other languages or domains.
    
\end{abstract}

\section{Introduction}

Speech conveys various types of information, where affective (``emotional") content is one aspect that has a role in social interactions. The way how affect is realized in speech appears to have certain universal properties \cite{prinz2004emotions, Russell_crosscultural, aff_indonesian}, but there are language- and culture-specific aspects as well \cite{Wierzbicka_fundamental_emotions, Evans_language_universals_myth}. This also means that the study of affective expressions and their subjective interpretation calls for language-specific data resources. However, up to this date, there has not been a suitable speech corpus to study expression and perception of affect in spontaneous Finnish, limiting the extent that affect has been studied in Finnish from linguistic and technological perspectives. 

This paper describes our sample selection process for creating the first corpus of affective spontaneous Finnish based on three existing large-scale speech corpora. We specifically emphasize our approach for informed but automatic selection of affect-rich samples for manual annotation, a process we refer to as \textit{affective speech mining}. We describe how we applied affective speech mining, how the data were annotated, and provide a description of the resulting corpus. We also analyze to what extent our affect mining approach yielded more diverse affective speech samples compared to a random sampling approach. Finally, with the help of the obtained affect annotations, we conduct post-hoc analyses to investigate which methodological choices in the affective mining process would have yielded the most diverse samples for annotation. 

As a result, the paper introduces a new and the first corpus for affective spontaneous speech in Finnish and provides speech scientists with methodological findings on how to effectively choose samples for manual annotation from datasets that are too large for manual analysis.

\subsection{Prior work}

In the context of speech technology, the analysis of affective speech is related to automatic speech emotion recognition (SER), where a suitable corpus or corpora are required for training and testing a SER system. There are a few existing Finnish corpora with labels for emotional content, such as \cite{FESC}, \cite{VAARAS20239}, and \cite{mediateam}. However, these corpora are relatively small in scale, and consist of acted speech \cite{FESC,mediateam} or are from a very specific communicative setting \cite{VAARAS20239}. This limits their representativeness in terms of real-world spontaneous speech \cite{real_vs_acted_speech, schuller_cross_corpus_ser, ser_sota_and_lessons_learnt_from_the_first_challenge, LSSED}.

More generally, a large proportion of existing SER corpora of different languages also consist of acted speech (e.g., \cite{schuller_cross_corpus_ser, ser_sota_and_lessons_learnt_from_the_first_challenge, LSSED, Waaramaa_nonsense_sentences, EMODB, RAVDESS}), as acting is a practical way to obtain large amounts of speech data with known emotional labels. Spontaneous emotional speech corpora are much fewer \cite{emo_corpus_review}, as post-hoc annotation of affective content is time-consuming and finding expression-rich speech samples from natural conversational data is not trivial. For instance, a notable proportion of spontaneous everyday speech is likely to be relatively neutral in terms of affective content, especially in certain cultures, thereby also constituting the majority of data if random sampling is used to select data for annotation.

Given that typical annotation efforts are limited to a small subset of all available speech data in the given language, there is a need for efficient methods for automatic discovery of speech samples that would cover a broad range of affective expression in the given domain. This is especially important for the development of SER systems, where coverage of different phenomena is often more important than representativeness of the sample (which can usually be obtained with random sampling).

To facilitate effective annotation efforts, various techniques for \textit{informed} sample selection have been developed, often falling under the umbrella of \textit{active learning} (AL). In AL, the idea is to use properties of the data, potentially paired with a machine learning model trained on an annotated subset of the data, to reduce human annotation effort by automatically selecting samples for annotation following some set criteria. In the context of SER, various AL approaches have been successfully used to streamline the data annotation process. These include uncertainty- and diversity-based methods \cite{al_dimensional_SER, al_incremental_adaptation, al_ser_dnn, al_uncertainty_sampling}, crowdsourced annotations with trustability-weighted agreements \cite{trustability_al}, and clustering-based methods like $k$-medoids clustering \cite{VAARAS20239}, which have been used to identify key speech samples for manual annotation of emotional content. Overall, these studies highlight the benefit of informed sample selection to minimize annotation efforts. In the present study, a clustering-based AL approach is used for affect mining from a corpus which is too large to be annotated manually.

\section{Creation of the FinnAffect corpus}

The aim of FinnAffect creation was to obtain manual annotations for a broad range of affective phenomena present in Finnish spontaneous speech. For this purpose, we utilized several source corpora with somewhat distinct communicative contexts and speaker populations, and combined them with our affect mining approach to select speech samples for manual annotation.  

\subsection{Data sources}

We used three large-scale spontaneous Finnish speech corpora as the source for the FinnAffect corpus: Lahjoita Puhetta (``LP"; \cite{LP}), TamPuhe (``TP"; \cite{tampuhe_en}), and HelPuhe (``HP"; \cite{helpuhe1_en}). LP is a large-scale speech corpus collected as part of a nationwide speech audio crowd-sourcing project consisting of 3,270 hours (1,687 hours transcribed) of spontaneous speech samples from over 20,000 native Finnish speakers. In LP, the speakers were prompted with various everyday topics and were asked to tell about them freely (see \cite{LP}). The TP and HP datasets are significantly smaller than the LP dataset, consisting of interview recordings from the 1970's, 1990's and 2010's. The original purpose of the recordings was the study of the over-time change in regional dialects of the Tampere and Helsinki areas of Finland. 

All three sources of data were first compiled into a single pool of utterances. All original source corpora contain text transcriptions, but with differing temporal alignments. LP transcriptions come with word level alignments, whereas in TP and HP the unit of alignment varies from one word to several clauses. For TP and HP, all clips shorter than 20 s were treated as individual utterances. Transcripts of longer turns in HP and TP were force-aligned at the word level using the Aalto ASR tool \cite{aalto_ASR}, after which 300-ms silence threshold was applied to segment the clips in LP, HP, and TP into utterances. Finally, all utterances from 1- to 20-s in duration and with an automatically estimated speech-to-noise ratio (SNR) of 20 dB or higher (using the tool from \cite{brouhaha}) were maintained for the sample selection phase. This resulted in a total of 1,474,728 utterances (1,438,537 utterances from LP, 26,407 from TP, and 9,784 from HP). 

\subsection{Sample selection with affect mining} \label{subsec_sample_selection}

We targeted a total of 12,000 samples for annotation. To maximize the diversity of affective content, we chose to use medoid-based active learning (MAL) \cite{tuni_mal}, which is a clustering-based AL approach previously applied to sound event classification \cite{tuni_mal} and SER \cite{VAARAS20239}. In standard MAL, the idea is to first represent all samples in a feature space, and then apply $k$-medoids clustering with the same number of clusters as is the total sample target for annotation. By annotating the medoids, the assumption is that other samples in the cluster would potentially belong to the same ``class", and hence the medoid labels could be propagated to the rest of the cluster samples (see also \cite{VAARAS20239}). By initializing the clustering with farthest-first traversal (FAFT; \cite{faft_original}), a set of maximally distinct samples could be used as initial medoids in MAL, hoping to capture maximal data diversity in the process.

To represent our source data for MAL, we applied three types of features that we expected to be sensitive to (but not necessarily accurate of) affective content: 1) acoustic features using the eGeMAPS feature set designed for affective computing \cite{egemaps}, compressed down to 42-dim with PCA, 2) 6-dim cross-linguistic speech emotion classifier posterior features from the ExHuBERT system trained for SER on a diverse set of languages \cite{exhubert}, and 3) 3-dim text sentiment analysis posteriors (negative, neutral, positive) features obtained from the speech transcripts using the FinnSentiment \cite{finnsentiment} tool. Each eGeMAPS feature was z-score normalized at a speaker-level to reduce the impact of speaker-dependent variability. All three feature types were then normalized to have an equal contribution ($\frac{1}{3}$ each) to the total variance in  concatenated 51-dim feature vectors, hence balancing the contributions of each feature type in the distance calculations of FAFT and MAL.

Our initial aim was to use FAFT in the resulting feature space to find diverse initial cluster centroids, following \cite{VAARAS20239}. However, the memory requirements of standard $k$-medoids clustering grows quadratically with the number of data points, preventing its use on the present 1.4M samples. Also, the CLARA-approximation \cite{kaufman2009finding} of $k$-medoids clustering for large datasets does not support initialization with specific samples. Moreover, even with a powerful computing cluster, executing CLARA was not computationally feasible for our target annotation budget with nearly 10k clusters. Thereby, FAFT could not be used for cluster initialization, and CLARA with heuristic initialization (see \cite{sklearn}) was used to calculate 1,500 clusters for the data.

Given the clusters, 6 samples were then chosen from each cluster for annotation. These included the medoid and its five closest samples with the constraint that two samples from each source corpora had to be chosen. In cases with less than 2 samples per source corpus in a cluster, the missing samples were randomly chosen from all the clusters to achieve the target count. As a result, a total of 9,000 samples (1500 x 6) were obtained with the approach (4,028 from LP, 2,577 from TP, and 2,395 from HP). In addition, we chose 3,000 samples randomly (1,000 per data source) to enable comparison to the affect mining approach and to obtain a representative distribution of affective content in the data, resulting in a total of 12,000 samples to annotate.

\subsection{Annotation process}

Five native Finnish listeners (3 female, age 20--30 years; self-reported normal hearing) were hired to annotate the samples, receiving a financial compensation for the task. Each annotator was tasked with 4,000 samples (2,000 common for all annotators). In addition, 20 \textit{quality assurance} samples were annotated three times by each annotator, batched so that five always appeared three times in random positions within a batch of 1,015 subsequent samples. This was done to ensure within-annotator consistency during the annotation process.

The annotators conducted the annotation in a self-paced manner across one month of time using a Python-based software with a graphical user interface (GUI) for the task, after an initial instruction session with the annotation tool. Samples were presented together with 4 s of preceding context with a clearly audible volume increase from the context to the actual target content. The annotators could listen to each sample a maximum of two times, and they were asked to rate their subjective perception of emotional arousal (``\textit{virittyneisyys}" in Finnish) and valence (``\textit{tunnesävy}") using two continuous sliders on a scale -1.0 to 1.0 with endpoints marked as ``low" vs. ``high" for arousal and ``negative" vs. ``positive" for valence. The annotators could annotate as many samples as they wanted per day, but they had to have at least a 15-min break after every 30 min of annotation. All annotators delivered all the 4,000 annotations tasked to them.

We first z-score normalized all ratings from each annotator, as we found that the annotators varied in their use of the rating scale, and then scaled the result back to [-1.0,1.0] with max(abs()) across all the scores for easier interpretation. The normalization significantly improved the inter-annotator agreement on the 2,000 sample "\textit{gold standard}" (GS) set with five annotations (see below). For the present analyses, we calculated the mean valence and arousal ratings for each sample in the GS set and combined them with the 10,000 samples with single annotations. We derived discrete arousal and valence labels with thresholds of [low] $\le$ 0 $<$ [high] for arousal, and [negative] $\le$ $-$0.08 $<$ [neutral] $\le$ +0.08 $<$ [positive] for valence. The valence threshold was optimized to maximize pairwise Cohen's kappa between all the annotators and the majority vote on the GS set. As a result, the mean Spearman correlation (continuous scores) and Cohen's kappa (discrete labels) between each annotator and the mean/mode of annotators were $\rho = 0.85$ and $\kappa = 0.52$ for valence and $\rho = 0.89$ and $\kappa = 0.64$ for arousal on the GS samples, reflecting substantially above-chance agreement.

\section{Corpus description}

\begin{figure}[t]
\centering
\includegraphics[width=0.47\textwidth]{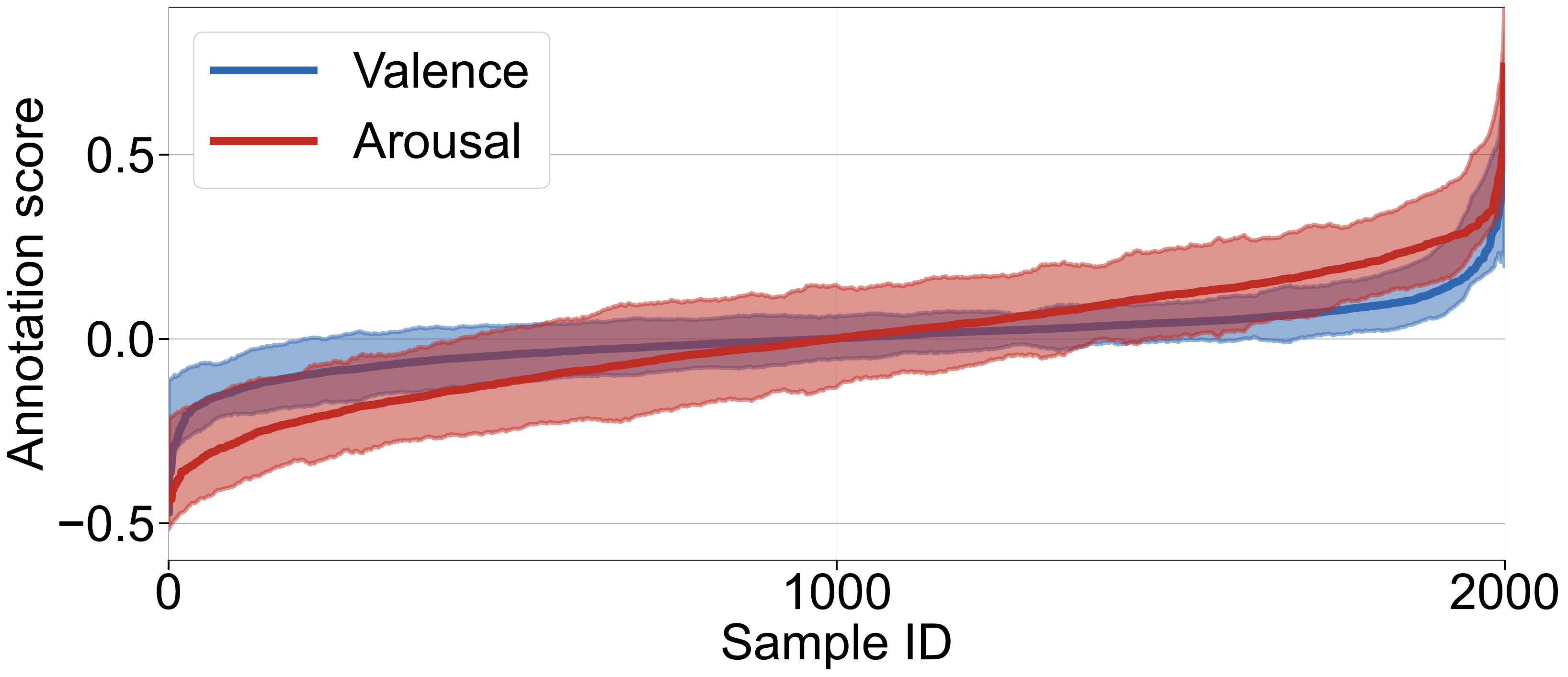}
\vspace{-11pt}
\caption{Annotated arousal and valence scores on the GS subset (2,000 samples with 5 annotations), sorted by the mean across annotators. Shading denotes $\pm1$ SD across the annotators.}
\label{fig:valence_arousal_sorted}
\end{figure}

Table \ref{table:corpus_sample_specifics} summarizes the resulting corpus properties. Fig.\ref{fig:valence_arousal_sorted} shows the valence and arousal score distributions on the GS set and the variance across the annotators. The overall distributions were similar for the 10k single-annotation samples and are not shown separately. As can be observed, there is a systematic distribution of low, medium, and high scores for both valence and arousal. After discretization, there is a relatively balanced set of low vs. high arousal classes, whereas most valence labels are neutral, but still with thousands of samples for both negative and positive valence. In addition, the data has a large number of speakers (N = 3936) and samples from both genders. 

\begin{table}[t]
\centering
\vspace{-6pt}
\caption{Summary statistics of the resulting FinnAffect corpus.}
\vspace{-9pt}
\includegraphics[width=0.46\textwidth]{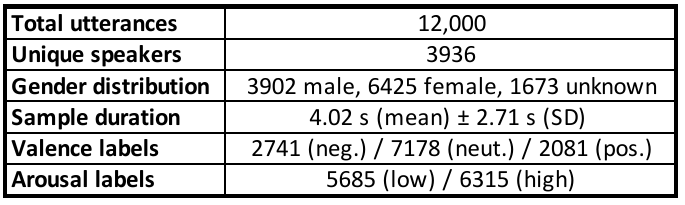}
\label{table:corpus_sample_specifics}
\vspace{-12pt}
\end{table}

To test if the affect mining strategy yielded more diverse samples compared to the random sampling, we compared the variances of the sample annotation scores in the two sets with Levene's test \cite{levenes_test}. There were no statistical differences between the annotator scores for either of the affect dimensions ($p > 0.05$ for both). The corresponding class label proportions were also highly similar for both cases, indicating that the diversity of annotation scores for the samples chosen with the affect mining strategy did not differ from those selected with random sampling. 

\section{Post-hoc analysis of affect mining}

Given that the affect mining did not result in more diverse emotional contents compared to random sampling, we analyzed the reasons for this finding and whether some alternative choices would have resulted in a more diverse set of samples. One hypothesis was that perhaps the MAL-based approach simply converges to a similar outcome with random sampling with a sufficiently large number of samples (clusters) to annotate. Alternatively, we speculated that the lack of FAFT initialization in the CLARA clustering might have caused the cluster density to follow the overall density distribution of the data, thereby aligning with random sampling from the same distribution.  

To study these questions, we used the 12,000 annotated samples to simulate the sample selection process in a post-hoc manner. We operationalized sample diversity as the standard deviation (SD) of the valence/arousal scores in a given selected subsample of data points, assuming that larger variation in the scores reflects richer variation in affective expression. We calculated the diversities for increasing sample sizes sampled from the 12k samples, comparing a) random sampling, b) samples obtained from FAFT, c) $k$-medoid medoids after clustering the data with the FAFT initialization, and d) $k$-medoids from the CLARA algorithm. For each dataset size, random sampling diversity scores were averaged across 100 runs, whereas FAFT and clustering scores were averaged across 5 runs of the test to average out stochastic variation in the results. For comparison, we ran CLARA using both $k$++ and heuristic initializations in Scikit-learn, and also CLARA in MATLAB using the $k$++ init.
    
\begin{figure*}[t]
\centering
\includegraphics[width=0.98\textwidth]{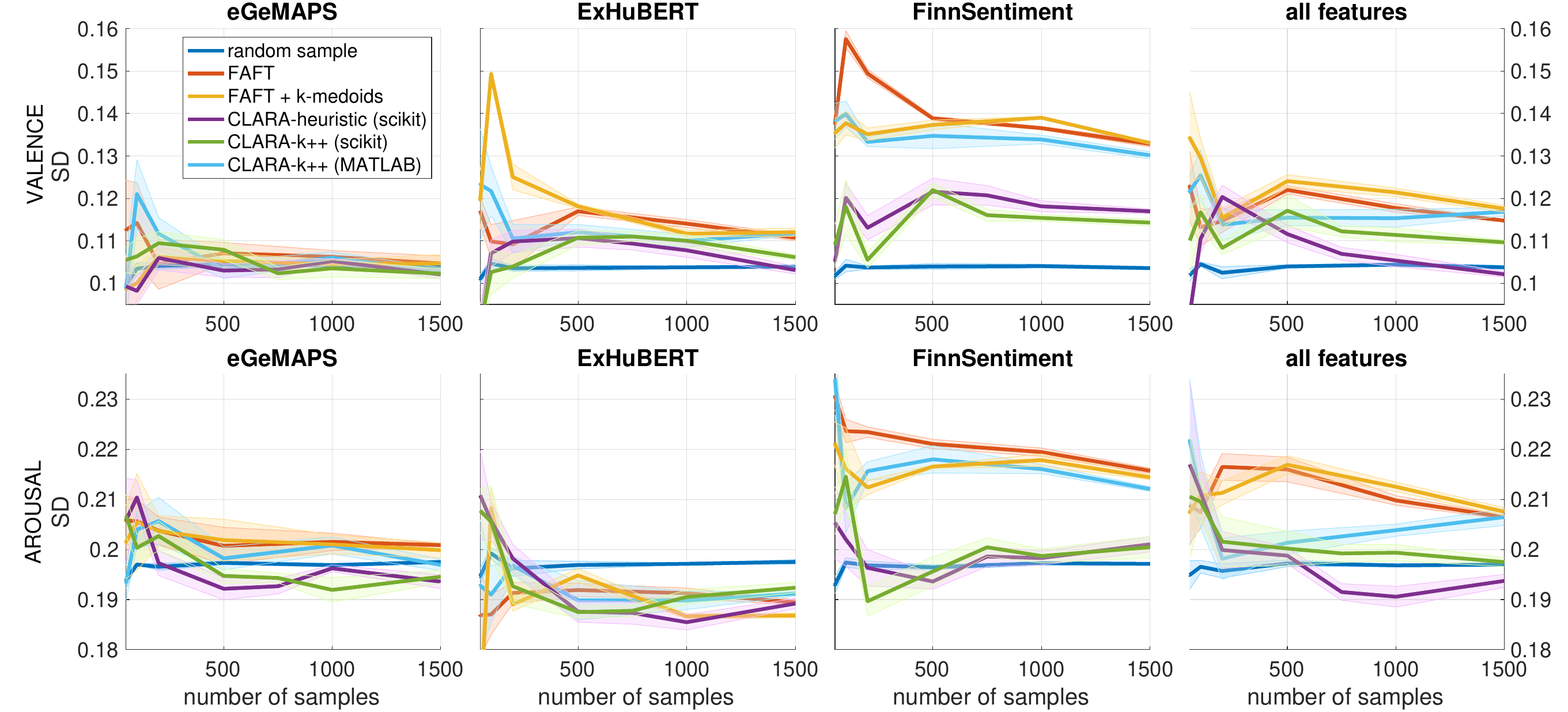}
\caption{Standard deviation (SD) of annotator scores for valence (top) and arousal (bottom) in case of different simulated sample selection strategies for annotation, and as a function of number of samples chosen. Shading denotes $\pm1$ standard error across different runs/samplings.}
\label{fig:STD_scores}
\end{figure*}

Fig. \ref{fig:STD_scores} shows the results of the analysis. As can be observed, samples from FAFT, FAFT-initialized k-medoids, and MATLAB CLARA without FAFT all result in higher affective diversity (annotation score SD) compared to random sampling at least up to 1,500 samples. Both FAFT and FAFT + $k$-medoids also outperform CLARA at all but the smallest and the very largest sample sizes, demonstrating the general benefit of FAFT in the discovery of diverse samples from data. The MATLAB CLARA implementation with $k$-medoids++ initialization yielded a greater diversity of annotation scores than both Scikit-learn implementations. The reason for this difference in the implementations is not clear. 

Importantly, the heuristic initialization in Scikit-learn CLARA, as also used in the original sample selection for annotation of FinnAffect, resulted in comparable results to random sampling at 1500 clusters, which explains why the original clustering-based sampling diversity did not differ significantly from random sampling. The Scikit-learn CLARA clustering with $k$-medoids++ initialization outperforms the heuristic initialization, but still does not outperform the MATLAB implementation. While all the methods start to gradually converge towards random sampling with increasing sample/cluster counts, there is still a clear margin to random sampling at the maximum of 1500 clusters for all but the Scikit-learn heuristic CLARA approach when using the full feature set. 

As for the post-hoc analysis of features, eGeMAPS does not appear to be useful for affective content mining in the given corpus, whereas ExHuBERT is beneficial for valence and sentiment scores appear to help for both valence and arousal. 

We also conducted post-hoc comparison of different clustering algorithms and distance metrics using the 12,000 annotated samples to test if some configuration results in a higher cluster purity (better organization of data into clusters of similar affect). Purity was calculated as the average proportion of the most frequent class in each cluster, excluding single-sample clusters that artificially increase apparent cluster purity. We experimented with the $k$-means, bisecting $k$-means, agglomerative, and CLARA clustering algorithms. For agglomerative clustering, we experimented with the Euclidean, Manhattan, cosine, and Chebyshev distance metrics. These distance metrics were also tested with CLARA with the addition of Pearson correlation distance. Furthermore, CLARA was tested with both the heuristic and $k$-medoids++ initializations, resulting in a total of 16 different clustering algorithm variants. We ran each experiment using either the eGeMAPS, ExHuBERT, or FinnSentiment features, and also with all features combined into 51-dim feature vectors (see Sec. \ref{subsec_sample_selection}). For each algorithm variation, we tested $k \in \{ 50, 60, ..., 250, 500, 750, 1000, 1500 \}$ clusters and repeated each experiment 10 times to account for the randomness of the cluster initialization process, resulting in a total of 16,000 cluster purity experiments.

\begin{figure}[t]
\centering
\vspace{-6 pt}
\includegraphics[width=0.385\textwidth]{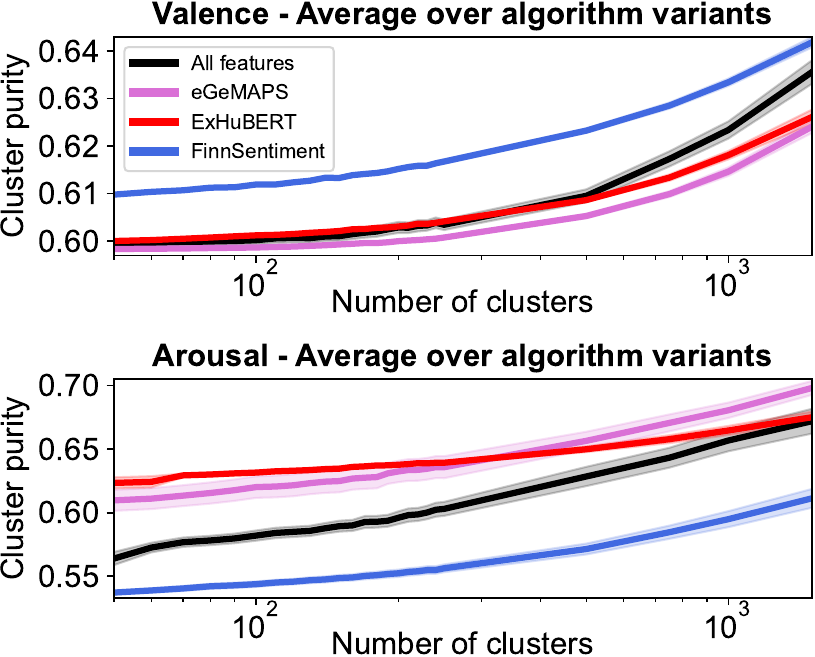}
\caption{The results of the cluster purity experiments for the four feature types, averaged across all 16 clustering algorithm variants (standard error as shaded area). The results are shown for valence (top) and arousal (bottom).}
\label{fig:cluster_purity_average_over_algorithms}
\vspace{-14pt}
\end{figure}


Fig. \ref{fig:cluster_purity_average_over_algorithms} presents the results of the cluster purity experiments for different features. The text-based FinnSentiment features cluster more systematically than the other features in case of valence, but are notably inferior for arousal, highlighting the strong link between valence in speech and its content. When averaged over both valence and arousal, eGeMAPS and ExHuBERT exhibit superior purity for all cluster counts, but with a high number of clusters, the combination of all features become similar. In terms of the clustering algorithms (not shown separately), all performed similarly except for agglomerative clustering, where only Euclidean distance yielded results comparable to the other variants. On average, CLARA was the best clustering algorithm regardless of its hyperparameters, followed by $k$-means, bisecting $k$-means, and agglomerative clustering (Mann-Whitney \textit{U} test (MWUT) \cite{mann_whitney_u_test_original}, $p < 0.05$ for all pairs). When further inspecting the different hyperparameter configurations for CLARA using MWUT, there were no significant differences between heuristic or $k$-medoids++ initialization ($p > 0.05$) when averaging across all other factors. Overall, the best distance metrics for CLARA were cosine and Pearson correlation (no significant differences), followed by Manhattan, Euclidean, and Chebyshev distances (all $p < 0.05$ compared to cosine and Pearson correlation, and with each other). All in all, the cluster purity analyses reflect a complementary perspective to the affect mining process, revealing how well the affective content is organized within the clusters in the representation space. This contrasts with the sample selection analyses, which measure the extent to which different clusters represent different emotional contents.

\section{Conclusions}

This paper presented the FinnAffect corpus for studying affect expression and developing SER systems for spontaneous Finnish. Moreover, we introduced a pipeline for mining affect-rich speech samples for manual annotation, making use of a rich set of speech descriptors from prosodic, cross-linguistic speech emotion, and text sentiment perspectives. Although the affect mining approach did not result in more diverse human labels compared to random sampling in our dataset creation, a series of post-hoc analyses revealed that the use of FAFT for sample discovery or $k$-medoids clustering initialization improves the resulting dataset diversity in terms of affective content. As a result, the use of either FAFT or FAFT-initialized $k$-medoids clustering, paired with affect-sensitive features, is recommended as a potential approach for affective content mining from large datasets.

\section{Acknowledgements}
The work of KL was funded by the CONVERGENCE project, a grant awarded by the Jane and Aatos Erkko Foundation to Tampere University. The authors wish to thank the annotators for their efforts. The authors would also like to thank Tampere Center for Scientific Computing for the computational resources used in this study.

\bibliographystyle{IEEEtran}
\bibliography{mybib}

\end{document}